\documentclass[12pt,fleqn]{article}

\usepackage{multirow}

\usepackage{microtype,amsbsy}
\usepackage{graphicx}
\usepackage{subcaption}
\usepackage{booktabs} 
\usepackage[percent]{overpic}

\usepackage{times,enumitem}
\usepackage{hyperref}

\usepackage{amsmath}
\usepackage{amssymb,dsfont}
\usepackage{mathtools}
\usepackage{amsthm}

\usepackage[capitalize,noabbrev]{cleveref}
\usepackage{times,enumitem}

\usepackage{graphicx}

\usepackage[labelfont=bf]{caption}

\usepackage[square,numbers]{natbib}
\usepackage{booktabs}

\usepackage[top=1in,bottom=1in,left=1in,right=1in]{geometry}

\usepackage{hyperref}

\usepackage{xcolor}

\usepackage{wrapfig}

\usepackage[font=footnotesize]{caption}

\usepackage{amsfonts}
\usepackage{amsmath}

\usepackage{parskip}

\usepackage[compact]{titlesec}

\usepackage{soul}

\usepackage{helvet}
\usepackage{sectsty}
\allsectionsfont{\sffamily}

\usepackage{fancyhdr}
\renewcommand{\headrulewidth}{0.4pt}

\usepackage[highstructure]{accessibility}
\theoremstyle{plain}

\theoremstyle{definition}

\theoremstyle{remark}

\begin{document}
\title{FAST: Factorizable Attention for Speeding up Transformers}
\author{Armin Gerami$^*$, Monte Hoover$^*$, Pranav S. Dulepet, Ramani Duraiswami\\ Perceptual Interfaces and Reality Laboratory, \\ Department of Computer Science, University of Maryland, College Park, USA}
\date{\tt\{agerami, mhoover4, pdulepet, ramanid\}@umd.edu}

\vskip 0.3in



\maketitle
\begin{abstract}
	Motivated by the factorization inherent in the original fast multipole method and the improved fast Gauss transform we introduce a factorable form of attention that operates efficiently in high dimensions. This approach reduces the computational and memory complexity of the attention mechanism in transformers from $O(N^2)$ to $O(N)$. In comparison to previous attempts, our work presents a linearly scaled attention mechanism that maintains the full representation of the attention matrix without compromising on sparsification and incorporates the all-to-all relationship between tokens. We explore the properties of our new attention metric and conduct tests in various standard settings. Results indicate that our attention mechanism has a robust performance and holds significant promise for diverse applications where self-attention is used.
\end{abstract}
\section{Introduction}
\label{introduction}
Transformers are the single deep learning architecture that underpin many recent successful applications in diverse fields, including in natural language processing, speech, computer vision, and biology. 


Transformers incorporate a Softmax-based all-to-all score computation mechanism denoted as ``attention". While this mechanism has proven to be extremely effective in learning tasks, they have a cost that is quadratic in the length of the input ($N$) and in the data dimension ($D$), and need a similar amount of memory. Our goal is to develop a more efficient transformer implementation that is as expressive using a novel attention formulation described in \S~\ref{method}. 


\subsection{Increasing $N$: Long Range attention}
Transformers are becoming the architecture of choice due to their ability to model arbitrary dependencies among tokens in a constant number of layers. Training is now done on more and more tokens and data to make models more expressive and learn more complex relationships. Due to the quadratic dependence on $N$, the problem of ``long range attention'' that arises needs to be tackled, and is dealt with in the literature in various ways.

\begin{enumerate}[leftmargin=0.25cm]
	\item {\bf Algorithmic:} Faster attention algorithms via various approaches, including spectral matrix decompositions, kernel method approximations, and sparsification via locality sensitive hashing have been proposed ~\cite{dingLongNetScalingTransformers2023,wangLinformerSelfattentionLinear2020,pmlr-v119-katharopoulos20a, kitaevReformerEfficientTransformer2020,jiangMistral7B2023,beltagy2020longformer}, but these solutions have not appeared to have gained traction. This appears to be for the most part due to perceived lower expressivity of the resulting transformers. These algorithms are further discussed in \S \ref{related-works}. 
	\item {\bf Parallelization: }Efficient attention via careful parallelization and minimizing communication between the CPU and the GPU \cite{dao2022flashattention, dao2023flashattention}. This is widely used, but is still quadratic. There are also quadratic frameworks to extend training across multiple nodes and GPUs to allow larger problems to fit in memory \cite{Rasley2020-dg, Singh2022-wx}.
	\item {\bf Non-transformer Models:} New learning architectures are being proposed as an alternative to Transformers. Recent ones which have generated considerable interest include Mamba~\cite{guMambaLinearTimeSequence2023}, Retentive networks~\cite{sunRetentiveNetworkSuccessor2023}, and CRATE~\cite{Yu2023-bk}.These do require re-engineering the model pipeline. 
\end{enumerate}

In practice, the really large models use quadratic vanilla attention strategies, and  work on the data in batches at various stride-lengths and stitch together the results to reach token lengths to the trillions \cite{Chowdhery2023-bf, touvron2023llama}. Previously proposed fast attention mechanisms do not appear to have been integrated into these works. 

\subsection{Contributions of this paper}
\par  We present FAST, a novel algorithm that achieves $O(N)$ computational and memory complexity for calculating an attention-based score without compromising accuracy or sparsifying the matrix. FAST is based on a new class of attention metrics, which are factorizable (in the spirit of the fast multipole method \cite{GreengardRokhlin1987} and the improved fast Gauss transform \cite{Yang2003-mg}) to achieve a linear cost. The computations are simple and support automatic differentiation. Our implementation of FAST can be seamlessly integrated in any Transformer architecture, or wherever there is need for an  attention mechanism. 

To evaluate our algorithm, we perform various tests\vspace{-0.2cm}
\begin{itemize}[leftmargin=0.25cm]
	\setlength{\itemsep}{-0.1cm}
	\item We compare the cost of performing forward pass against Softmax for various values of $N$ and $D$.
	\item We analyze the resulting attention matrix on the basic datasets, such as MNIST \cite{Lecun1998-ww}.
	\item To measure the expressivity of our algorithm, we compare our performance against Softmax on the five tasks comprising the Long Range Arena~\cite{tayLongRangeArena2020}. 
\end{itemize}

\section{FAST}
\label{method}
We describe below the computation of the self attention score in transformers using both the ``vanilla" Softmax and our novel attention metric, Fastmax. 
\par \textbf{Notation:} Bolded upper case letters, e.g., $\mathbf{X}$ indicate matrices, and bolded lower case letters $\mathbf{x}_i$ and $\mathbf{x}_{ij}$ the $i$th row and the element at the $i$th row and $j$th column of $\mathbf{X}$ respectively. Unbolded letters $X,x,x_i,x^{(i)}_{jkl}$ indicate scalars.
\subsection{Preliminaries}
\label{preliminary}
In vanilla attention, given a sequence of $N$ tokens, channel dimension of $C$, $H$ heads, and channel per head dimension of $D = C/H$, each head takes in three trainable matrices $\mathbf{Q},\mathbf{K},\mathbf{V} \in \mathbb{R}^{N\times D}$, and gives an output score matrix $\mathbf{O} \in \mathbb{R}^{N\times D}$. Score calculation is formulated using a matrix vector product (MVP) of the attention $\mathbf{A}$ with $\mathbf{V}$ as follows:
\begin{align}
	\mathbf{O} &= \mathbf{A}\mathbf{V},\quad \mathbf{A} = \mbox{Softmax}\left(\mathbf{Q}\mathbf{K}^T\right)\\
	\mathbf{o_{ij}} &= \dfrac{\sum_{n=1}^{N}{\exp(\mathbf{q}_i^T\mathbf{k_n}/\sqrt{D})\mathbf{v_{nj}}}}{\sum_{n=1}^{N}{\exp(\mathbf{q}_i^T\mathbf{k}_n/\sqrt{D})}}.\label{0}
\end{align}

Each head requires $O(N^2D)$ computations and memory. The final score is given by concatenating scores from each head, with a total $O(N^2C)$  computation and memory. We can apply a causal mask by changing Eq.~\ref{0} to
\begin{align}
	\mathbf{O} &= \mbox{tril}(\mathbf{A}) \mathbf{V}, \quad \mbox{tril}(\mathbf{A})_{ij} = \begin{cases}\mathbf{a}_{ij}, j\leq i\\0,\, j > i\end{cases};\\
	\mathbf{o}_{ij} &= \dfrac{\sum_{n=1}^{i}{\exp(\mathbf{q}_i^T\mathbf{k}_n/\sqrt{D})\mathbf{v}_{nj}}}{\sum_{n=1}^{i}{\exp(\mathbf{\hat{q}}_i^T\mathbf{k_n}/\sqrt{D})}}.
\end{align}
\par We should mention that the similarity metric of $\mbox{exp}(q\cdot k)$ used in vanilla attention is not the only metric deriving an attention matrix, but rather the common approach. Any convex function can substitute for the similarity metric, but the expressiveness of the model hinges on the chosen function. During training, the weight matrices creating the Query, Key, and Value matrices are updated. 

\subsection{Proposed attention Metric}
We first normalize $\mathbf{q}$ and $\mathbf{k}$ and use the polynomial kernel $f(x)$ as our similarity metric for deriving $\mathbf{A}$, that is
\begin{align}
	\mathbf{\tilde{q}}_i &= \mathbf{q}_i - \mbox{mean}(\mathbf{q}_i),\quad\mathbf{\tilde{k}}_i = \mathbf{k}_i - \mbox{mean}(\mathbf{k}_i),\label{6}\\
	\mathbf{\hat{q}}_i &= \mathbf{\tilde{q}}_i/\mbox{STD}(\mathbf{\tilde{q}}_i),\quad \mathbf{\hat{k}}_i =\mathbf{\tilde{k}}_i/\mbox{STD}(\mathbf{\tilde{k}}_i),\label{7}\\
	\mathbf{a}_{ij} &= \dfrac{f(\mathbf{\hat{q}}_i^T\mathbf{\hat{k}}_j)}{\sum_{n=1}^{N}f(\mathbf{\hat{q}}_i^T\mathbf{\hat{k}}_n)},\\
	f(x) &= \sum_{\ell=0}^p \dfrac{x^\ell}{\ell !},\quad f:\mathbb{R}\rightarrow\mathbb{R},\label{taylor}
\end{align}
which is inspired by the Taylor expansion of $e^{x}$, though the resulting attention matrix is {\em distinct from the exponential}, and is valid as long as Eq.~\ref{valid} is satisfied. Due to its relationship with the exponential, the approach provides an attention mechanism that is as robust as the original. We refer to this process as Fastmax; i.e.,
\begin{align}
	\mathbf{A} &= \mbox{Fastmax}(\mathbf{QK}^T)\\
	\mathbf{a}_{ij}&\geq 0, \quad \sum_{i=0}^N  \mathbf{a}_{ij} = 1\label{valid}
\end{align}

The score for each head is given by the MVP
\begin{align}
	\mathbf{O} = \mathbf{A}\mathbf{V};
\end{align}
which can be broken down to
\begin{align}
	\mathbf{o}_{ij} &= \dfrac{\sum_{n=1}^{N}{f(\mathbf{\hat{q}}_i^T\mathbf{\hat{k}_n} )\mathbf{v}_{nj}}}{\sum_{n=1}^{N}{f(\mathbf{\hat{q}}_i^T\mathbf{\hat{k}_n} )}}.\label{tobeexpanded}
\end{align}
We continue this section assuming that $p = 2$ in Eq. \ref{taylor}; i.e., $f(x) = 1 + x + x^2/2$, which contains the results for $p=1$ as well. We  provide the results for $p = 1, 2$ in Section~\ref{results}.

\subsection{Gradient Bound}
It has been suggested \cite{Qin2022-mp} that the lack of expressivity in other subquadratic attention mechanisms rises due to the possibility of their gradients either blowing up or vanishing. To assess the robustness of our proposed solution, we will now examine the behavior of the gradient of the Score, $\nabla \mathbf{o}_{ij}$, and show that it is lower and upper bounded. Let us denote $\mathbf{\hat{q}}_{i}.\mathbf{\hat{k}}_{j}$ as $s_{ij}$, and write the attention and Score as
\begin{align}
	\mathbf{o}_{ij} &= \sum_{n=1}^{N}{\mathbf{a}_{in}\mathbf{v}_{nj}},\quad \mathbf{a}_{in} = \dfrac{f(s_{ij})}{\sum_{m=1}^{N}f(s_{im})}.
\end{align}
Taking the derivative with respect to $s_{il}$
\begin{align}
	\dfrac{\partial\mathbf{a}_{in}}{\partial s_{il}} &= \begin{cases}\dfrac{1+s_{il}}{\sum_{m=1}^{N}f(s_{im})}(1-\mathbf{a}_{ij}),& n = l\vspace{5pt
		}\\\dfrac{1+s_{il}}{\sum_{m=1}^{N}f(s_{im})}(-\mathbf{a}_{ij}),& n\neq l\end{cases}\\
	\dfrac{\partial\mathbf{o}_{ij}}{\partial s_{il}} &= \sum_{n=1}^{N}\dfrac{\partial\mathbf{a}_{in}}{\partial s_{il}}\mathbf{v}_{nj} = \dfrac{1+s_{il}}{\sum_{m=1}^{N}f(s_{im})}(\mathbf{v}_{lj} - \sum_{n=1}^{N}\mathbf{a}_{in}\mathbf{v}_{nj}).\label{bound}
\end{align}
Since $0\leq s_{il}, \mathbf{a}_{in} \leq 1$, the first term in Eq.~\ref{bound} is bounded within $0\leq \dfrac{1+s_{il}}{\sum_{m=1}^{N}f(s_{im})} \leq \dfrac{5}{2N+3}$, and the second term $0\leq \mathbf{v}_{lj} - \sum_{n=1}^{N}\mathbf{a}_{in}\mathbf{v}_{nj} \leq 2\left \lVert \mathbf{v}^T_{j} \right \rVert_\infty$, where $\mathbf{v}^T_{j}$ is the $j$th collumn of $\mathbf{V}$. Therefore $\dfrac{\partial\mathbf{o}_{ij}}{\partial s_{il}}$ is bounded within $0\leq \dfrac{\partial\mathbf{o}_{ij}}{\partial s_{il}} \leq \dfrac{10\left \lVert \mathbf{v}^T_{j} \right \rVert_\infty}{2N+3}$. During backpropagation in neural networks, excessively large gradients can lead to disastrous exploding/vanishing updates when propagated through activation functions. Our attention mechanism mitigates such risks by establishing a firm upper bound, ensuring stability and smooth learning.

\subsection{Factorization}
\label{Implementation}
We will use multidimensional Taylor series to provide a factorization for the exponential in $D$ dimensions. This closely follows the method developed for high dimensional Gaussians in \cite{Yang2003-mg}, but is adapted to the case of exponentials. Note that this allows us to develop a valid attention metric that satisfies Eq \ref{valid}, has similar near-field and far-field behavior to the exponential softmax, and is faster to compute.

First, to understand why factorization speedsup computation, we present a simple example. Consider the MVP 
\begin{equation}
	\begin{bmatrix}a_1b_1& a_1b_2& a_1b_3\\ a_2b_1& a_2b_2& a_2b_3 \\ a_3b_1& a_3b_2& a_3b_3\end{bmatrix} \times \begin{bmatrix}d_1\\ d_2 \\ d_3 \end{bmatrix} = \begin{bmatrix}u_1\\ u_2 \\ u_3\end{bmatrix}.\label{simpmult}
\end{equation}
The naive method of calculating $u$ would be
\begin{align}
	u_i = a_ib_1d_1 + a_ib_2d_2 + a_ib_3d_3,
\end{align}
with a total of $9$ multiplications and $6$ accumulations. Applying factorization, we have
\begin{align}
	x &= b_1d_1 + b_2d_2 + b_3d_3,\quad
	u_i = a_ix,
\end{align}
reducing the operations to $6$ multiplications and $2$ accumulations. For an $M\times M$ matrix-vector multiplication with the same structure as in Eq.~\ref{simpmult}, the number of operations reduces from $O(M^2)$ to $O(M)$ by applying factorization.
The score calculation  in Eq.~\ref{tobeexpanded} can be broken down to matrix multiplications in the form of Eq.~\ref{simpmult}. Specifically,
\begin{align}
	\mathbf{o}_{ij} &= \dfrac{\mathbf{f}_{ij}}{\mathbf{g}_i},\quad \mathbf{F} \in \mathbb{R}^{N\times D}, \mathbf{G} \in \mathbb{R}^{N}\label{bigo}
\end{align}
\begin{align}
\!\!\!\!\!\!\!\!\!\!\!\! \mathbf{F} \!= \!\left(\mathbf{I} \!+\! \sum_{m=0}^{D}\!{\begin{bmatrix}\mathbf{\hat{q}}_{1m}\mathbf{\hat{k}}_{1m}& \dots& \mathbf{\hat{q}}_{1m}\mathbf{\hat{k}}_{Nm}\\ \vdots& \ddots& \vdots \\ \mathbf{\hat{q}}_{Nm}\mathbf{\hat{k}}_{1m}& \dots& \mathbf{\hat{q}}_{Nm}\mathbf{\hat{k}}_{Nm}\end{bmatrix}}\!+\! \sum_{m,l=0}^{D}\!\begin{bmatrix}\mathbf{\hat{q}}_{1m}\mathbf{\hat{k}}_{1m}\mathbf{\hat{q}}_{1l}\mathbf{\hat{k}}_{1l}& \dots& \mathbf{\hat{q}}_{1m}\mathbf{\hat{k}}_{Nm}\mathbf{\hat{q}}_{1l}\mathbf{\hat{k}}_{Nl}\\ \vdots& \ddots& \vdots \\ \mathbf{\hat{q}}_{Nm}\mathbf{\hat{k}}_{1m}\mathbf{\hat{q}}_{Nl}\mathbf{\hat{k}}_{1l}& \dots& \mathbf{\hat{q}}_{Ni}\mathbf{\hat{k}}_{Ni}\mathbf{\hat{q}}_{Nj}\mathbf{\hat{k}}_{Nj}\end{bmatrix}  \right)\!\mathbf{V},\label{bigf}
\end{align}
\begin{align}
\!\!\!\!\!\!\!\!\!\!\!\! \mathbf{G}= \left(\mathbf{I} + \sum_{m=0}^{D}{\begin{bmatrix}\mathbf{\hat{q}}_{1m}\mathbf{\hat{k}}_{1m}& \dots& \mathbf{\hat{q}}_{1m}\mathbf{\hat{k}}_{Nm}\\ \vdots& \ddots& \vdots \\ \mathbf{\hat{q}}_{Nm}\mathbf{\hat{k}}_{1m}& \dots& \mathbf{\hat{q}}_{Nm}\mathbf{\hat{k}}_{Nm}\end{bmatrix}}+ 
	\sum_{m,l=0}^{D}\begin{bmatrix}\mathbf{\hat{q}}_{1m}\mathbf{\hat{k}}_{1m}\mathbf{\hat{q}}_{1l}\mathbf{\hat{k}}_{1l}& \dots& \mathbf{\hat{q}}_{1m}\mathbf{\hat{k}}_{Nm}\mathbf{\hat{q}}_{1l}\mathbf{\hat{k}}_{Nl}\\ \vdots& \ddots& \vdots \\ \mathbf{\hat{q}}_{Nm}\mathbf{\hat{k}}_{1m}\mathbf{\hat{q}}_{Nl}\mathbf{\hat{k}}_{1l}& \dots& \mathbf{\hat{q}}_{Ni}\mathbf{\hat{k}}_{Ni}\mathbf{\hat{q}}_{Nj}\mathbf{\hat{k}}_{Nj}\end{bmatrix} \right) \mathds{1}\label{bigg}
\end{align}
where $\mathds{1}\in\mathbb{R}^N$ is a vector of all ones; i.e., 
\begin{align}
	\!\!    \mathbf{f}_{ij} \!&= \!\sum_{n=1}^{N}\!\left(1 + \!\sum_{m=1}^{D} \!{\mathbf{\hat{q}}_{im}\mathbf{\hat{k}}_{nm}} +\!\sum_{m,l=1}^{D}\!{\mathbf{\hat{q}}_{im}\mathbf{\hat{k}}_{nm}\mathbf{\hat{q}}_{il}\mathbf{\hat{k}}_{nl}}\right)\mathbf{v}_{nj},\label{fij}\\
	\!\!   \mathbf{g}_{i} \!&= \!\sum_{n=1}^{N}\left(1 + \sum_{m=1}^{D}\!{\mathbf{\hat{q}}_{im}\mathbf{\hat{k}}_{nm}} +\!\sum_{m,l=1}^{D}\!{\mathbf{\hat{q}}_{im}\mathbf{\hat{k}}_{nm}\mathbf{\hat{q}}_{il}\mathbf{\hat{k}}_{nl}}\right).\label{gi}
\end{align}
Changing the summation orders we get
\begin{align}
	\mathbf{f}_{ij} &= \sum_{n=1}^{N}{\mathbf{v}_{nj}} + \sum_{m=1}^{D}{\sum_{n=1}^{N}{\mathbf{\hat{q}}_{im}\mathbf{\hat{k}}_{nm}\mathbf{v}_{nj}}} +
	\sum_{m,l=1}^{D}{\sum_{n=1}^{N}{\mathbf{\hat{q}}_{im}\mathbf{\hat{k}}_{nm}\mathbf{\hat{q}}_{il}\mathbf{\hat{k}}_{nl}}\mathbf{v}_{nj}},\label{f}\\
	\mathbf{g}_{i} &= \sum_{n=1}^{N}{1} + \sum_{m=1}^{D}{\sum_{n=1}^{N}{\mathbf{\hat{q}}_{im}\mathbf{\hat{k}}_{nm}}} +
	\sum_{m,l=1}^{D}{\sum_{n=1}^{N}{\mathbf{\hat{q}}_{im}\mathbf{\hat{k}}_{nm}\mathbf{\hat{q}}_{il}\mathbf{\hat{k}}_{nl}}}.\label{g}
\end{align}
Applying factorization we get
\begin{align}
	\mathbf{f}_{ij} &= x^{(1)}_j + \sum_{m=1}^{D}{\mathbf{\hat{q}}_{im}x^{(2)}_{jm}} + \sum_{m,l=1}^{D}{\mathbf{\hat{q}}_{im}\mathbf{\hat{q}}_{il}x^{(3)}_{jml}},\\
	\mathbf{g}_{i} &= y^{(1)} + \sum_{m=1}^{D}{\mathbf{\hat{q}}_{im}y^{(2)}_{m}} + \sum_{m,l=1}^{D}{\mathbf{\hat{q}}_{im}\mathbf{\hat{q}}_{il}y^{(3)}_{ml}},
\end{align}
where,
\begin{align}
	x^{(1)}_j &= \sum_{n=1}^{N}\mathbf{v}_{nj}, \quad &x^{(2)}_{jm} = \sum_{n=1}^{N}\mathbf{\hat{k}}_{nm}\mathbf{v}_{nj},\quad
	&x^{(3)}_{jml} = \sum_{n=1}^{N}\mathbf{\hat{k}}_{nm}\mathbf{\hat{k}}_{nl}\mathbf{v}_{nj},\label{x}\\
	y^{(1)} &= N, \quad &y^{(2)}_{m} = \sum_{n=1}^{N}\mathbf{\hat{k}}_{nm},\quad
	&y^{(3)}_{ml} = \sum_{n=1}^{N}\mathbf{\hat{k}}_{nm}\mathbf{\hat{k}}_{nl}.\label{y}
\end{align}
\par Computing $x^{(1)},x^{(2)},x^{(3)}$ in Eq.~\ref{x} respectively require $O(ND)$, $O(ND^2)$, $O(ND^3)$ computation and $O(D)$, $O(D^2)$, $O(D^3)$ memory. Computing and $y^{(1)},y^{(2)},y^{(3)}$ in Eq.~\ref{y} respectively require $O(1)$, $O(ND)$, $O(ND^2)$ computation and $O(1)$, $O(D)$, $O(D^2)$ memory. Computing $F,G$ in Eq.s~\ref{f},\ref{g} require $O(ND^3)$, $O(ND^2)$ computations and $O(ND)$, $O(N)$ memory. In total, as written, FAST has a computational complexity of $O(ND^3)$ and memory complexity of $O(ND^2+D^3)$ per head, and $O(NH(C/H)^3)$ and $O(NH(C/H)^2 + H(C/H)^3)$ computational and memory complexity for all of the heads. Note that by increasing both number of heads $H$ and channels $C$, we can reduce the computational and memory cost; e.g., by quadrupling $H$ and doubling $C$, the computational cost halves.
\par To apply the causal mask, we change Eq.s~\ref{fij},\ref{gi} to 
\begin{align}
	\mathbf{f}_{ij} &= \sum_{n=1}^{i}(1 + \sum_{m=1}^{D}{\mathbf{\hat{q}}_{im}\mathbf{\hat{k}}_{nm}} +\sum_{m,l=1}^{D}{\mathbf{\hat{q}}_{im}\mathbf{\hat{k}}_{nm}\mathbf{\hat{q}}_{il}\mathbf{\hat{k}}_{nl}})\mathbf{v}_{nj},\\
	\mathbf{g}_{i} &= \sum_{n=1}^{i}(1 + \sum_{m=1}^{D}{\mathbf{\hat{q}}_{im}\mathbf{\hat{k}}_{nm}} +\sum_{m,l=1}^{D}{\mathbf{\hat{q}}_{im}\mathbf{\hat{k}}_{nm}\mathbf{\hat{q}}_{il}\mathbf{\hat{k}}_{nl}}),
\end{align}
where we changed the first summation range. Changing the summation orders and applying factorization we get
\begin{align}
	\mathbf{f}_{ij} &= x^{(1)}_j + \sum_{m=1}^{D}{\mathbf{\hat{q}}_{im}x^{(2)}_{jm}} + \sum_{m,l=1}^{D}{\mathbf{\hat{q}}_{im}\mathbf{\hat{q}}_{il}x^{(3)}_{jml}},\\
	\mathbf{g}_{i} &= y^{(1)} + \sum_{m=1}^{D}{\mathbf{\hat{q}}_{im}y^{(2)}_{m}} + \sum_{m,l=1}^{D}{\mathbf{\hat{q}}_{im}\mathbf{\hat{q}}_{il}y^{(3)}_{ml}},
\end{align}
where,
\begin{align}
	x^{(1)}_{1j} = \mathbf{v}_{1j},\quad x^{(1)}_{ij}& = x^{(1)}_{i-1j} + \mathbf{v}_{ij},\nonumber\\
	x^{(2)}_{1jm} =\mathbf{\hat{k}}_{1m}\mathbf{v}_{1j},\quad x^{(2)}_{ijm}& = x^{(2)}_{i-1jm} + \mathbf{\hat{k}}_{im}\mathbf{v}_{ij},\nonumber\\
	x^{(3)}_{1jml} =\mathbf{\hat{k}}_{1m}\mathbf{\hat{k}}_{1l}\mathbf{v}_{1j},\quad x^{(3)}_{ijml}& = x^{(3)}_{i-1jml} + \mathbf{\hat{k}}_{im}\mathbf{\hat{k}}_{il}\mathbf{v}_{ij},\label{xcausal}\\
	y^{(1)}_i = i, \quad y^{(2)}_{1m} = \mathbf{\hat{k}}_{1m}, \quad &y^{(2)}_{im} = y^{(2)}_{i-1m} + \mathbf{\hat{k}}_{im},\nonumber\\
	y^{(3)}_{1ml} = \mathbf{\hat{k}}_{1m}\mathbf{\hat{k}}_{1l}, \quad y^{(3)}_{iml}& = y^{(3)}_{i-1ml} + \mathbf{\hat{k}}_{im}\mathbf{\hat{k}}_{il}.\label{ycausal}
\end{align}
\par In the masked version of FAST, the factorized values are different for each row of $F$ and $G$, in contrast with the unmasked version's shared values. As a result, the memory complexity increase from $O(ND^2 + D^3)$ to $O(ND^3)$, while computational complexity remains $O(ND^3)$ per head. Considering all heads, the masked version's overall computational and memory complexity becomes $O(NH(C/H)^3)$. Similar to the unmasked version, we can reduce the cost by increasing both $H$ and $C$. In general, the costs in term of the control parameter $p$ becomes $O(NH(C/H)^{p+1})$ computational and $O(NH(C/H)^p + H(C/H)^{p+1})$ memory complexity for the unmasked Fastmax, and $O(NH(C/H)^{p+1})$ computational and memory complexity for the masked Fastmax.

\begin{figure}[htb!]
	\centering
	\includegraphics[width=0.66\columnwidth, trim=0mm 5mm 200mm 0mm, clip=true]{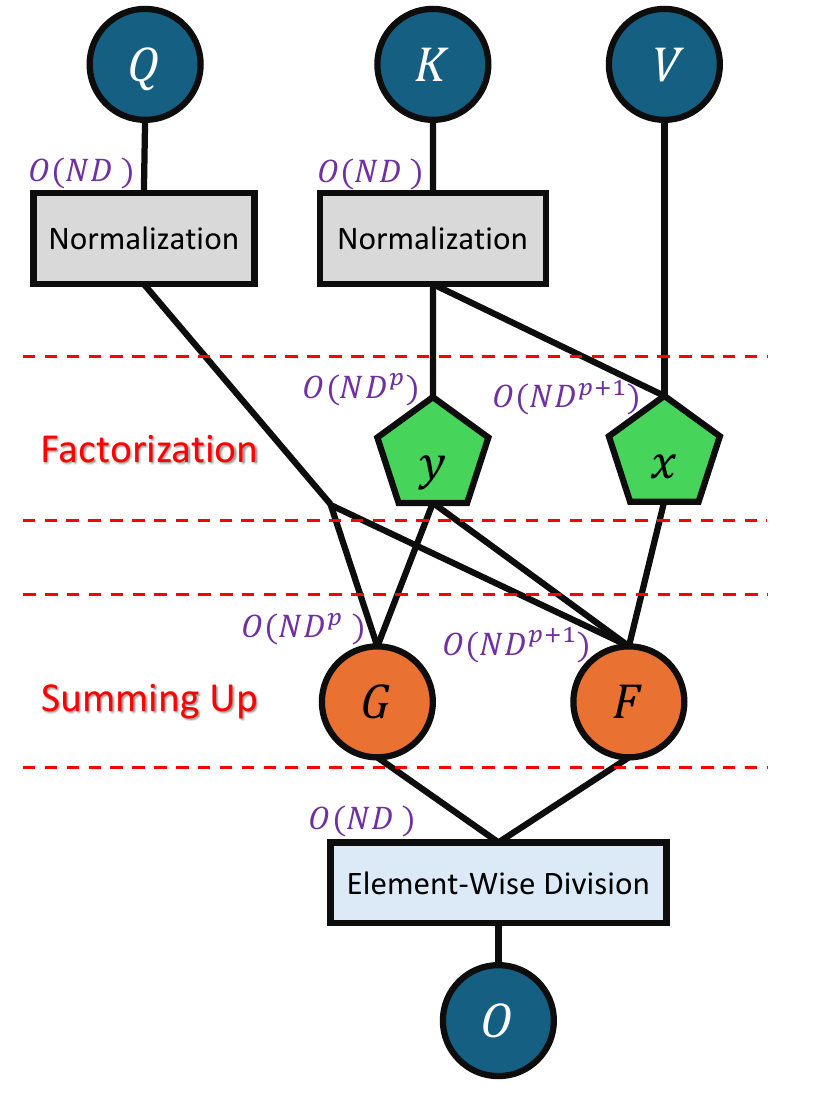}
	\caption{Flowchart of calculating Score using Fastmax. The purple terms on the upper-left of each step indicate their computational cost. The backward pass is computed using automatic differentiation, but further optimizations is possible using custom gradient (see \S \ref{custom_gradients}).}
	\label{fig:flow}
\end{figure}

\par Figure~\ref{fig:flow} shows the flowchart of our Score calculation process for a single head. Given $\mathbf{Q},\mathbf{K},\mathbf{V}$, the process starts with normalizing $\mathbf{Q}$ and $\mathbf{K}$ (Eq.s~\ref{6},\ref{7}) to get $\mathbf{\hat{Q}},\mathbf{\hat{K}}$. We then calculate the factorized values $x$ and $y$ using $\mathbf{\hat{K}}$ and $\mathbf{\hat{V}}$ (Eq.s~\ref{x},\ref{y}). Then, we find $\mathbf{F},\mathbf{G}$ using the factorized values and $\mathbf{\hat{Q}}$ (Eq.s~\ref{bigf},\ref{bigg}). The Score is given by the element-wise division $\mathbf{F}/\mathbf{G}$ (Eq.~\ref{bigo}).

\par One caveat to consider is the handling of dropout. When using Softmax, dropout is implemented by randomly omitting elements of the attention matrix. In the case of Fastmax, since the attention matrix is not explicitly created, dropout cannot be directly applied. Instead, dropout is applied to the factorized terms ($x$ and $y$ in Eq.s~\ref{x},\ref{y},\ref{xcausal},\ref{ycausal}). The choice of how to apply dropout to the factorized terms is not immediately obvious. One approach would be to dropout values uniformly from within the embedding dimensions of the factorized terms, or on the other end of the spectrum, dropout could be applied to all dimensions of a given $\mathbf{\hat{q}}_i$ or $\mathbf{\hat{k}}_j$ token before creating the factorized terms. 

\begin{figure}[htb!]
	\centering
	\includegraphics[width=0.8\columnwidth]{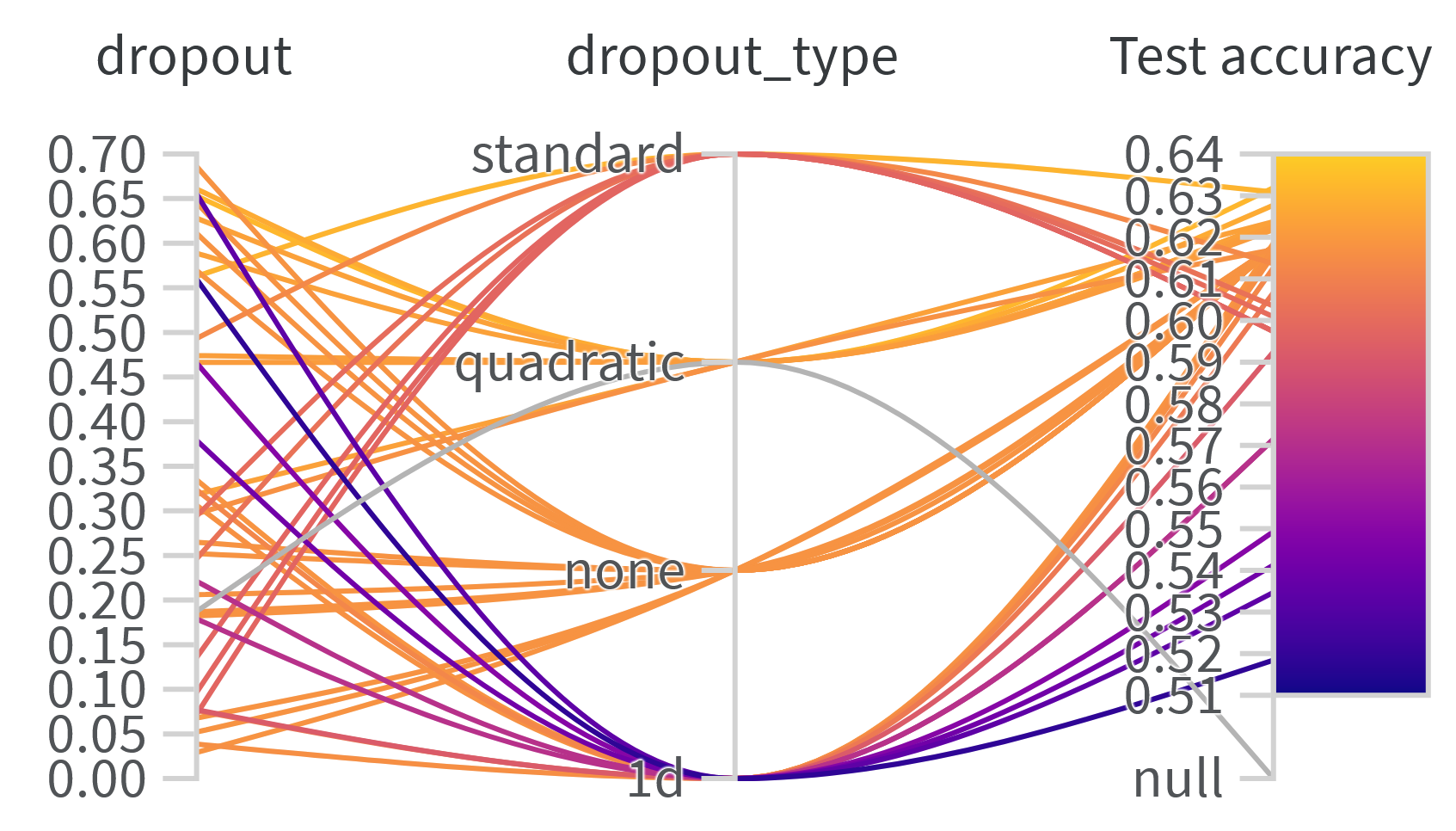}
	\caption{Empirical results of different dropout approaches. Notice that even small amounts of dropout on the quadratic term benefit test generalization.}
	\label{fig:dropout}
	\vspace{-8pt}
\end{figure}

Empirical results show a middle ground dropout approach to be the most effective. In Figure~\ref{fig:dropout} we show a comparison the approaches described above "standard" and "1d" respectively and an approach that only does dropout from within the embeddings of the quadratic terms of the factorization ("quadratic"). The quadratic approach proves to be the most effective, and the experiments also confirm the advantage even small amounts of this dropout provide over the alternative of omitting it entirely.

\subsection{Reducing Memory Cost with Custom Gradients}
\label{custom_gradients}
\par The memory cost of Fastmax can be reduced by implementing custom gradients. To elaborate, consider one head and $p = 2$. The memory cost dependency on $D^p$ for Fastmax arises from the necessity to store all the factorized terms to calculate the gradient update during the backward pass. Going back to Eq.~\ref{bound}, the derivative of the Score can be expressed as
\begin{align}
	\dfrac{\partial\mathbf{o}_{ij}}{\partial s_{il}} &= \dfrac{1+s_{il}}{\sum_{m=1}^{N}f(s_{im})}(\mathbf{v}_{lj} - \sum_{n=1}^{N}\mathbf{a}_{in}\mathbf{v}_{nj})\\
	& = \dfrac{1+\mathbf{\hat{q}}_{i}.\mathbf{\hat{k}}_{l}}{\sum_{m=1}^{N}f(\mathbf{\hat{q}}_{i}.\mathbf{\hat{k}}_{m})}(\mathbf{v}_{lj} - \dfrac{\sum_{n=1}^{N}f(\mathbf{\hat{q}}_{i}.\mathbf{\hat{k}}_{n})\mathbf{v}_{nj}}{\sum_{m=1}^{N}f(\mathbf{\hat{q}}_{i}.\mathbf{\hat{k}}_{m})}).
\end{align}

Put in to words, gradient of the Score function $\mathbf{O}$ can be calculated by storing $\mathbf{\hat{Q}}$, $\mathbf{\hat{K}}$, $\mathbf{V}$, $\sum_{m=1}^{N}f(\mathbf{\hat{q}}_{i}.\mathbf{\hat{k}}_{m})$ and $\sum_{n=1}^{N}f(\mathbf{\hat{q}}_{i}.\mathbf{\hat{k}}_{n})\mathbf{v}_{nj}$ for all $i$ and $j$ ($1\leq i \leq N$, $1\leq j \leq D$); a total of $O(ND)$ elements. Moreover, the latter two terms are already calculated during forward pass, and $\mathbf{\hat{q}}_{i}.\mathbf{\hat{k}}_{l}$ for all $i$ and $l$ can be calculated with $O(ND)$ computations using factorization, as explained earlier. The same goes for the masked version, with the difference that we should store $\sum_{m=1}^{i}f(\mathbf{\hat{q}}_{i}.\mathbf{\hat{k}}_{m})$ and $\sum_{n=1}^{i}f(\mathbf{\hat{q}}_{i}.\mathbf{\hat{k}}_{n})\mathbf{v}_{nj}$ instead of the latter two terms for all $i$ and $j$. In general, by using custom gradients, the computational complexity will remain $O(ND^{p+1})$, while the memory reduces to $O(ND^{p-1})$ for each head, for both masked and unmasked Fastmax.

\section{Results}
\label{results}
We have introduced the new Fastmax attention, demonstrated its linear scalability with the number of tokens, and its ability to yield stable gradient updates. In this section, we will show that our attention metric behaves as expected during implementation and that it is as expressive as Softmax. We will first compare Fastmax and Softmax as standalone blocks and then evaluate their performance on the Long Range Arena benchmark. Specifically, we will demonstrate the results for Fastmax with $p = 1$ and $2$, denoted as Fastmax1 and Fastmax2.

\begin{figure}[htb!]
	\centering
	\includegraphics[width=0.7\columnwidth, trim=0mm 0mm 250mm 0mm, clip=true]{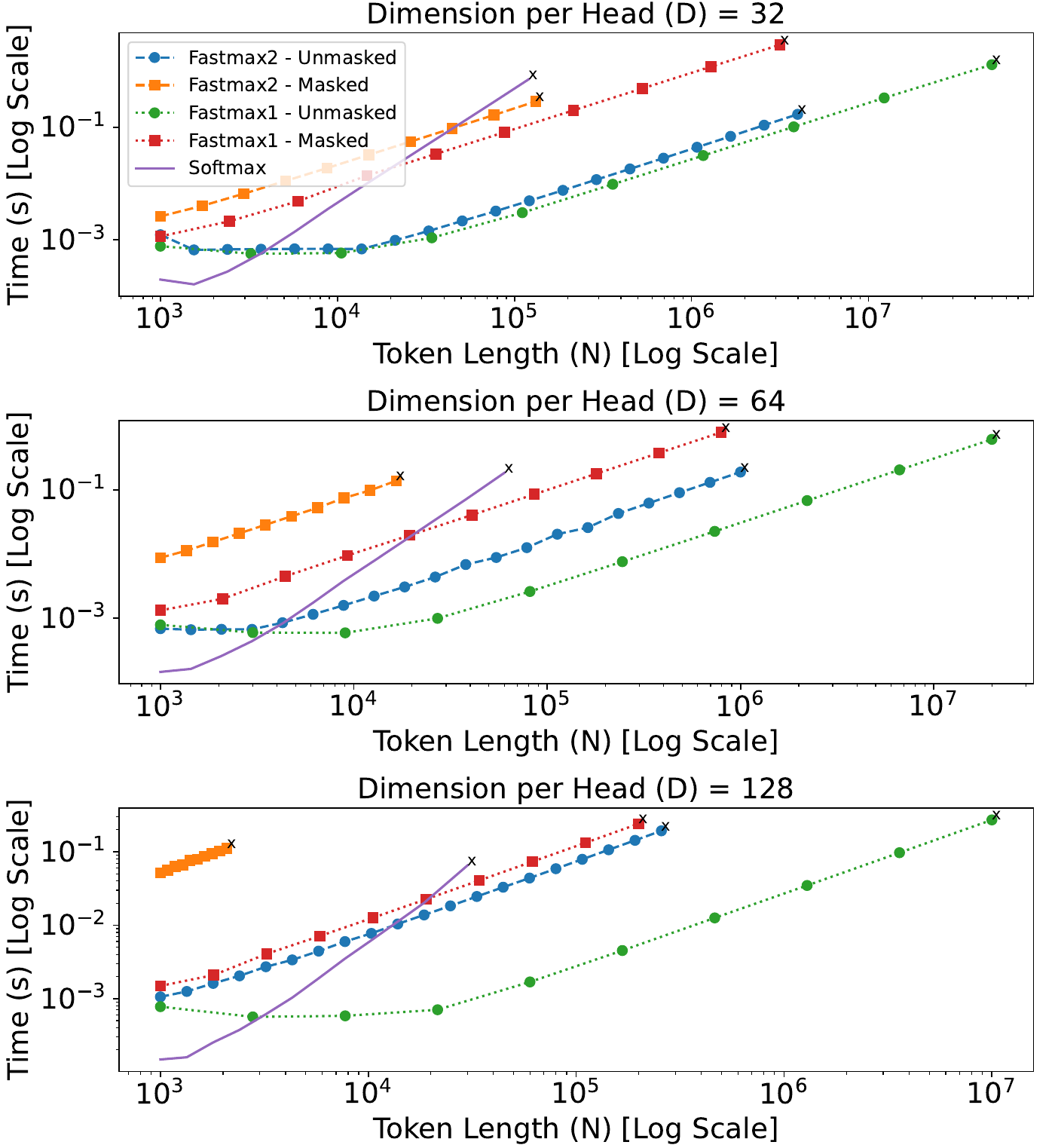}
	\caption{Comparison between times taken for calculating the Scores per head using Fastmax and Softmax on an RTX A6000 (48 GB memory) for various dimension per heads $D$. Softmax scales quadratically with number of tokens $N$, whereas Fastmax scales linearly. The 'x' marks indicate an ``out of memory" condition.}
	\label{fig:scaling}
\end{figure}


\subsection{Computational Efficiency}
Figure~\ref{fig:scaling} illustrates the wall-clock time for the forward pass of the Softmax, Fastmax1, and Fastmax2 for a single attention block on a log-log scale. As anticipated, the time for Fastmax scales linearly with $N$, while Softmax scales quadratically. Based on these wall-clock times, a model the size of Llama2\cite{touvron2023llama} with a head dimension $D=128$ gains speed and memory advantages with Fastmax1 at $N>1400$. Furthermore, we observe that the masked version of Fastmax has a $D\times$ higher wall-clock time than the unmasked version despite having the same computational cost. This is due to the increased memory consumption of incorporating the attention mask, which causes the GPU to serialize and thus reduces its parallelizability when using the same amount of memory. As explained in \S~\ref{custom_gradients}, there is an opportunity in future work to reduce the memory consumption by an order of $D$, subsequently bringing the down the wall-clock time further.

\subsection{Attention Map Visualizations}
\begin{figure}[htb!]
	\centering
	\begin{overpic}[width=0.9\columnwidth]{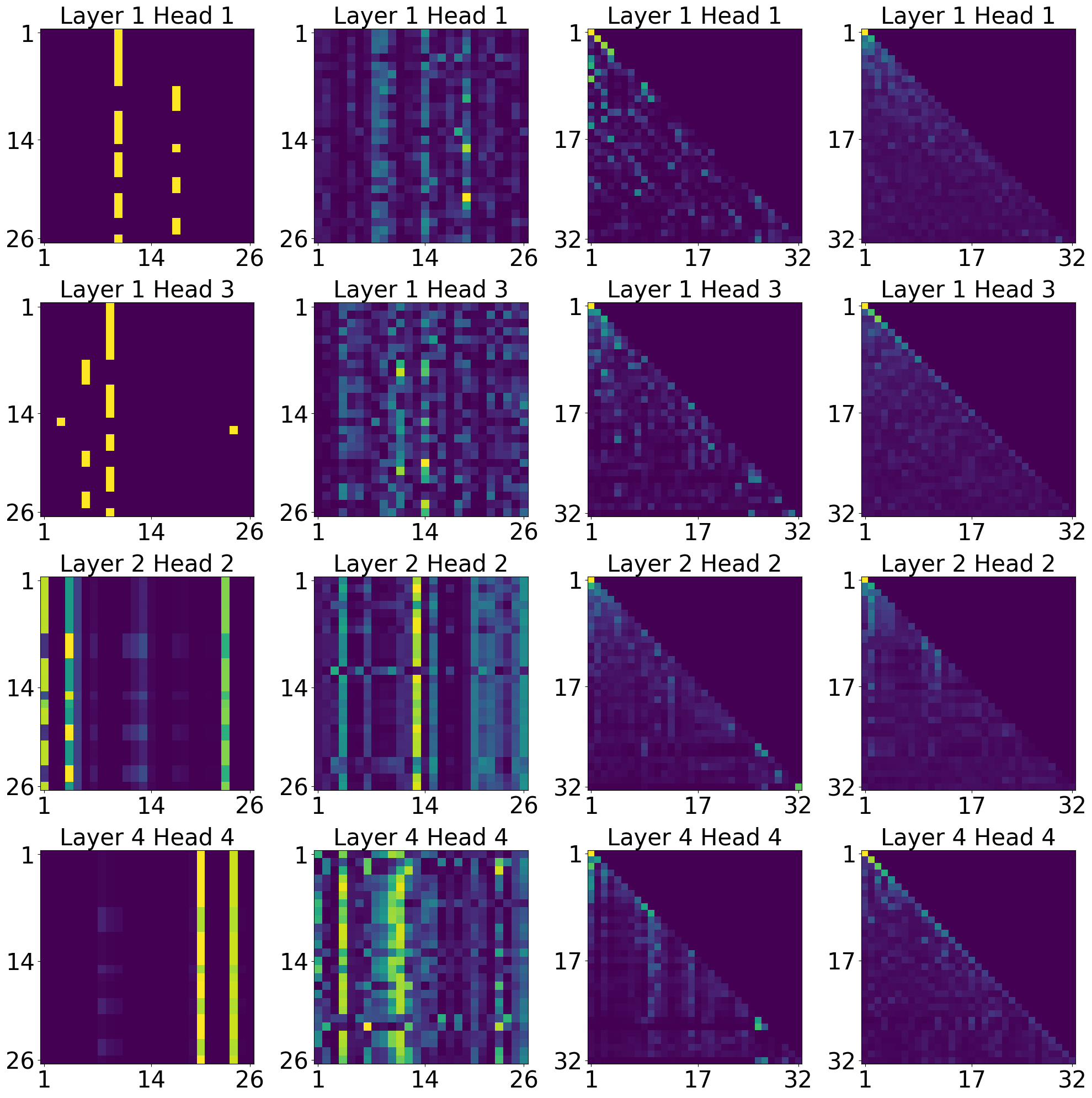}
		\put(11.5,-3.5){(a)}
		\put(36.5,-3.5){(b)}
		\put(61.5,-3.5){(c)}
		\put(86,-3.5){(d)}
	\end{overpic}
\vspace*{9pt}
	\caption{Attention maps from transformers trained on MNIST and Tiny Shakespeare data: (a) Softmax attention on MNIST, (b) Fastmax$2$ attention on MNIST, (c) Softmax attention on Tiny Shakespeare, (d) Fastmax attention on Tiny Shakespeare. Note that the mechanisms produce different scores, but both converge on training.}
	\label{fig:heat}
\end{figure}
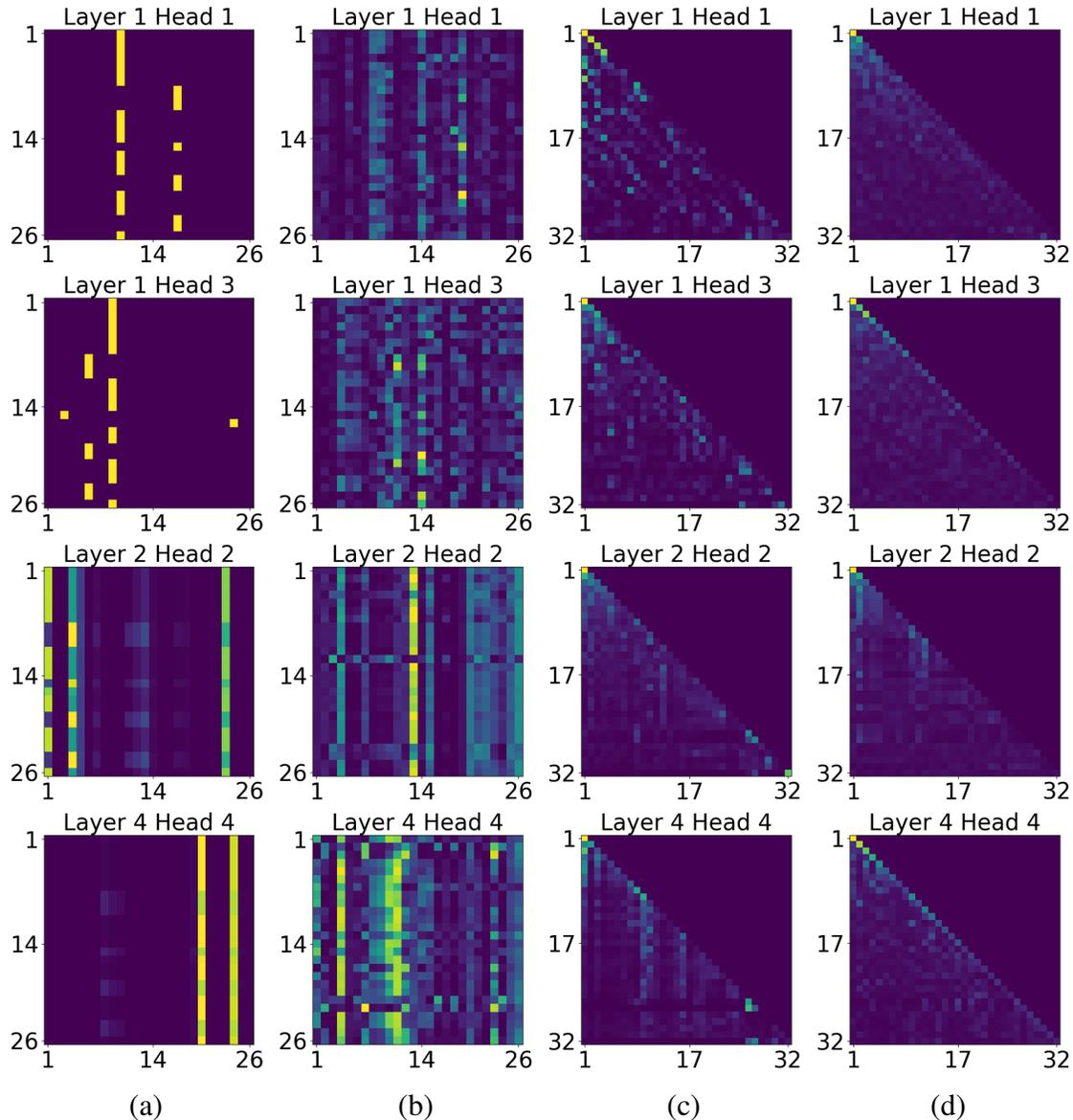

Figure~\ref{fig:heat} shows attention maps from randomly selected heads in the standard softmax transformer and from a Fastmax transformer, trained on the MNIST and the Tiny Shakespeare datasets. There are noticeable differences in the attention structure learned for image classification compared with text generation. The MNIST classifiers accumulate information from a small number of image patches with each attention head as indicated by the distinct columns, whereas the text generators maintain some amount of per-token information in each head, as indicated by the strong diagonal.  Fastmax maintains a structure recognizably similar to softmax attention, though there are some differences. We speculate that the similarity in structure indicates that the priors learned by a Fastmax transformer tend to be in line with those of a standard transformer, and this is further substantiated by the achieved results in the LRA test.

It's also worth noting that Fastmax attention is less localized than standard attention. Further investigation is required to determine whether a less localized attention has a positive or negative impact on performance. 

\subsection{Long Range Arena Results}
Earlier subquadratic attention formulations used the Long Range Arena (LRA) benchmark\cite{tayLongRangeArena2020} as a gauge of how capable a given formulation was at incorporating dependencies in long sequences and benchmarking the speed efficiency gained by the subquadratic attention. It turns out that high performance on the LRA benchmark often fails to transfer to other tasks such as next-token-prediction, as evidenced by the low perplexity scores of Performer\cite{choromanski2020rethinking} and Reformer \cite{kitaevReformerEfficientTransformer2020} on WikiText103 reported in \cite{poliHyenaHierarchyLarger2023}. Even so, LRA serves as a minimum bar for architectures to demonstrate long-range dependency capability, and it can illuminate the expressivity of a method by comparing performance within the various tasks. To better demonstrate the expressivity of Fastmax, and to gain a comprehensive understanding of its capabilities, we plan to conduct numerous additional tests on large datasets in our future work.

Table~\ref{tab:lra-acc} shows the achieved accuracies of various subquadratic transformer architectures on the five subtasks of the LRA benchmark. It is common for a subquadratic architecture to exceed the standard softmax dot product attention transformer on one task while falling noticeably behind in others, indicating that the architecture brings with it a set of different inductive priors (see Informer on the Pathfinder task and Linear Transformer on the ListOps task).  Fastmax in contrast is intended to exhibit the full expressivity of softmax dot product attention, including the priors and general expected behavior. The results on LRA seem to indicate that this behavior holds true in practice.

\begin{table*}[ht]
	\centering
	\caption{Long Range Arena results broken out by task. Scores represent accuracy on classification tasks. Note that a transformer with Fastmax attention maintains the full representation capabilities of standard softmax dot product attention. }
	\begin{tabular}{l | c c c c c |c}
		\toprule
		Model            & ListOps      & Text  & Retrieval   & Image & Pathfinder & Avg \\
		\hline
		Vanilla Trans.   & 38.37        & 61.95 & 80.69      & 40.57 & 65.26      & 57.37 \\
		\hline
		Informer~\cite{zhouInformerEfficientTransformer2020}         & 36.95& 63.60& 75.25& 37.55& 50.33&52.74\\
		Reformer~\cite{kitaevReformerEfficientTransformer2020}       & 37.00& 64.75& 78.50& 43.72& 66.40&58.07\\
		Linear Trans.~\cite{pmlr-v119-katharopoulos20a}    & 16.13& 65.90& 53.09& 42.34& 75.30&50.55\\
		Performer~\cite{choromanski2020rethinking}& 37.80& 64.39& 79.05& 39.78& 67.41&57.69\\
		\hline
		Fastmax2 (ours)& 37.40  & 64.30  & 78.11       & 43.18 & 66.55      & 57.90 \\
		Fastmax1 (ours)& 37.20  & 63.25  & 78.21       & 42.76 & 66.67      & 57.62 \\
		\bottomrule
	\end{tabular}
	\label{tab:lra-acc}
\end{table*}

\begin{table*}[ht]
	\centering
	\caption{Long Range Arena timing results, reported as training steps per second. Higher is better. All tests were performed on RTX A5000 GPUs using the same transformer architecture with head dimension $D=32$. Note that the theoretical break even point for second-order Fastmax with $D=32$ is $N=1024$, and these results confirm that in real training scenarios.}
	\begin{tabular}{l|ccccc|c}
		\toprule
		\multirow{2}{*}{Model}    & ListOps    & Text       & Retrieval  & Image      & Pathfinder &  Avg  \\
		& ($N=2000$) & ($N=4000$) & ($N=4000$) & ($N=1000$) & ($N=1000$) &       \\
		\hline
		Vanilla Trans.            &  6.4       &   1.8      &   1.7      &   3.0      &   6.1      &   3.8 \\
		\hline
		Fastmax$2$ (ours) & 11.6       &   6.1      &  6.9       &  3.0       &   6.8      &   6.9 \\
		Fastmax$1$ (ours) & 47.4       &  26.7      & 24.7       &  12.8      &  24.4      &  27.2 \\
		\bottomrule
	\end{tabular}
	\label{tab:lra-speed}
\end{table*}

\begin{figure}[htb!]
	\centering
	\begin{subfigure}{0.45\linewidth}
		\includegraphics[width=\linewidth]{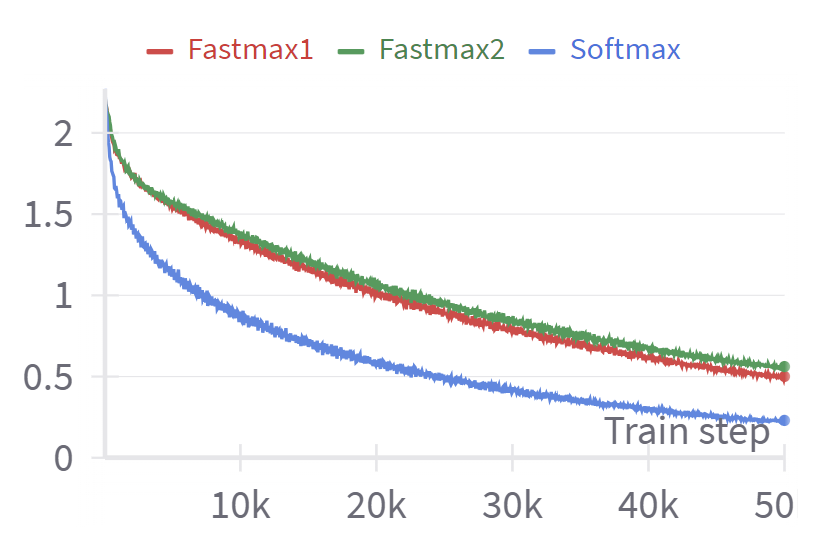}
		\caption*{(a)}
	\end{subfigure}
	\begin{subfigure}{0.45\linewidth}
		\includegraphics[width=\linewidth]{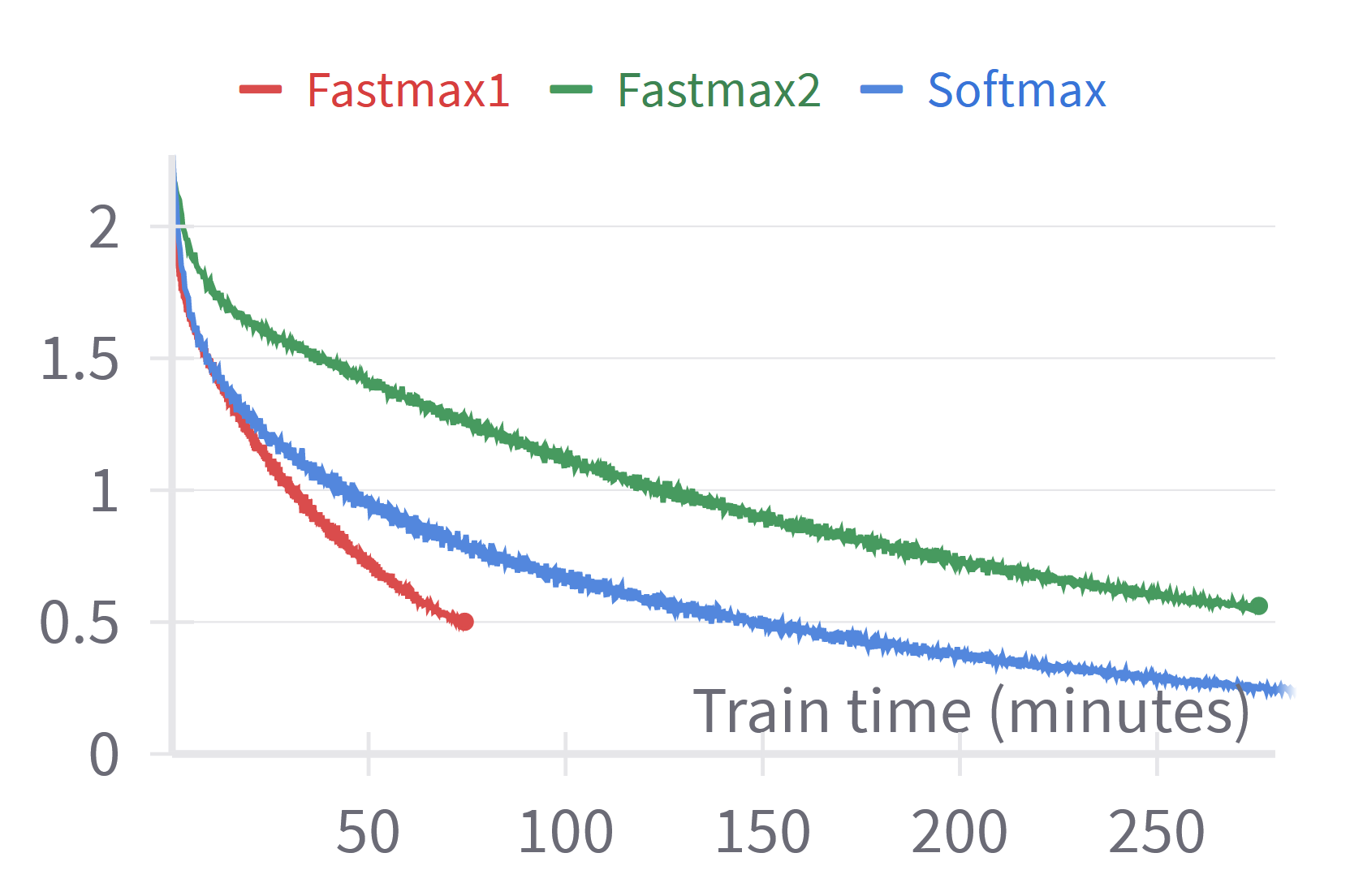}
		\caption*{(b)}
	\end{subfigure}
	\begin{subfigure}{0.45\linewidth}
		\includegraphics[width=\linewidth]{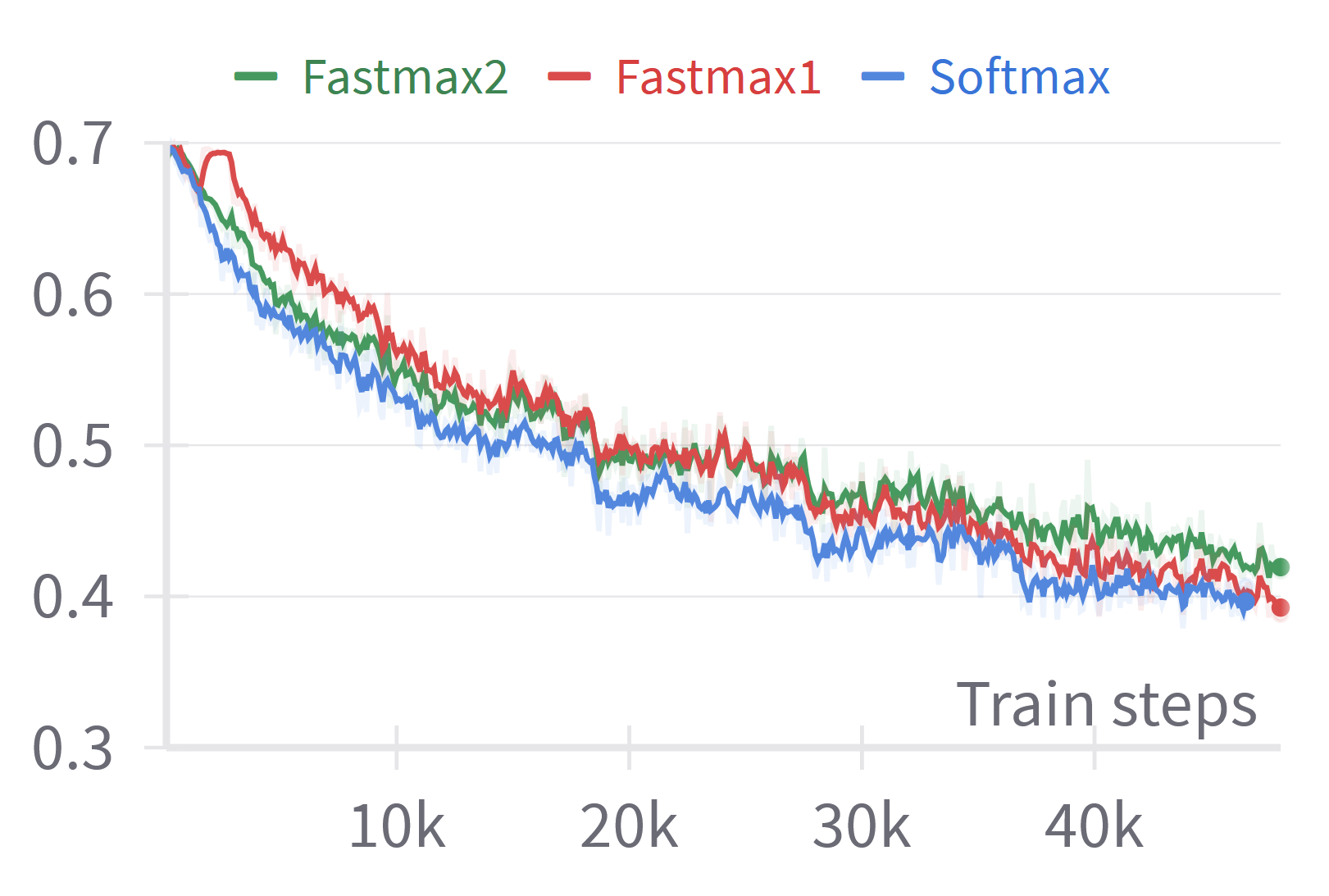}
		\caption*{(c)}
	\end{subfigure}
	\begin{subfigure}{0.45\linewidth}
		\includegraphics[width=\linewidth]{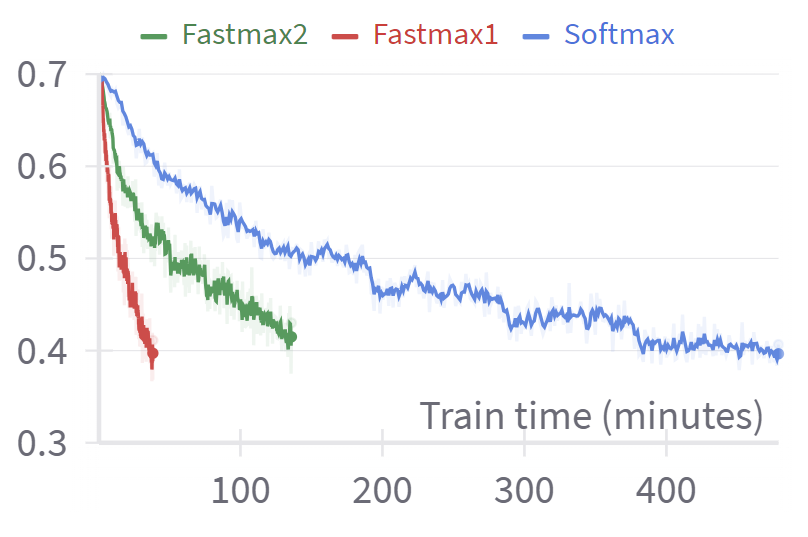}
		\caption*{(d)}
	\end{subfigure}
	\caption{Loss curves for training a standard softmax attention transformer compared with Fastmax1 and Fastmax2: (a) LRA Image loss plotted against the number of training steps, (b) LRA Image loss against wall clock time, (c) LRA Retrieval loss against the number of training steps, (d) LRA Retrieval loss against wall clock time. In the plots on the left we see Softmax converging at the same speed or faster than Fastmax1 when measured by train steps, but when measuring by wall clock time in the plots on the right we see Fastmax1 converging much faster.}
	\label{fig:loss-curves}
\end{figure}

The loss curves for different training scenarios offer another indication of softmax attention characteristics holding true in meaningful ways for Fastmax. In Figure~\ref{fig:loss-curves} we see that training loss curves for Fastmax generally follow the trajectory of softmax,
\begin{figure}[htb!]
	\centering
	\includegraphics[width=0.85\linewidth]{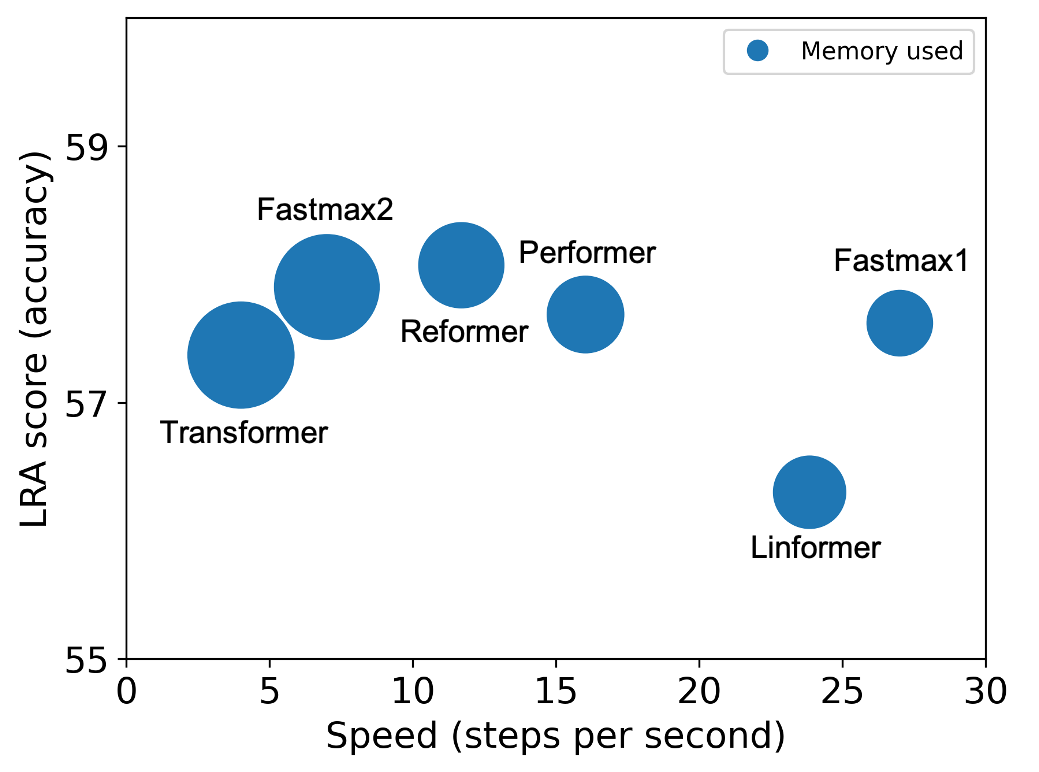}
	\caption{Long Range Arena results, after \cite{tayLongRangeArena2020}, showing speed versus accuracy for alternate transformer formulations, including the vanilla Transformer (softmax), and several others discussed in the paper, as well as the two Fastmax variants proposed in this paper. GPU memory usage is represented by circle area. The timings were measured on a RTX A5000 GPU. The hyperparameters for each algorithm were optimized. Further details are included in Tables \ref{tab:lra-acc} and \ref{tab:lra-speed}}
	\label{fig:lra}
\end{figure}

\section{Related Works}
\label{related-works}
Many sub-quadratic transformers have been proposed in the last few years and have been used occasionally in specific domains, but none so far have offered a viable alternative to softmax dot product attention in the general case. Reformer \cite{kitaevReformerEfficientTransformer2020} and Performer \cite{choromanski2020rethinking} both do well on the LRA benchmark, but have some drawbacks that make them unsuitable in some situations. Reformer achieves its efficiency through sparse attention, where the sparsity is selected through locality-sensitive hashing. This is quite effective but requires the same projection for the $\mathbf{Q}$ and $\mathbf{K}$ matrices, limiting expressivity. Performer is a kernel-based approach that approximates the attention matrix by using orthogonal random projections. In practice this has shown a drop in performance on large text datasets and in addition it requires an approximation of the $\mathbf{W^{(Q)}}$, $\mathbf{W^{(K)}}$, and $\mathbf{W^{(V)}}$ weight matrices, making it less suitable as a drop-in replacement for a model's attention layer. Linformer \cite{wangLinformerSelfattentionLinear2020} uses low rank decomposition of the attention matrix, and while it achieves the expected theoretical runtime efficiency in practice on GPUs, it also suffers the most in accuracy and expressivity. 

There has been a more recent wave of approaches that attempt to deal with some of these drawbacks. Flashattention \cite{dao2022flashattention, dao2023flashattention} has seen widespread adoption by tackling the problem of the gap between theoretical efficiency and actual wall clock time. The Flashattention CUDA kernels minimize CPU-GPU memory transfer, speeding up softmax attention by roughly 2x with no tradeoffs in expressivity. Because Fastmax is purely dot product-based, it can take advantage of Flashattention style GEMM kernels, and this will be a subject of future work. Recently transformer alternatives such as Mamba \cite{guMambaLinearTimeSequence2023} and CRATE \cite{yu2023whitebox} have shown promising results at efficiently extending to long context lengths but it remains to be seen how they perform at scale. Closest in spirit to this work is Hyena \cite{poliHyenaHierarchyLarger2023} which replaces attention with implicit convolutions and gating. Hyena also scales linearly and performs well on large language datasets, but does depart further from the characteristics of dense dot-product attention than Fastmax does. More work remains to be done on Mamba, CRATE, Hyena, and Fastmax to show that they definitively provide an advantage on difficult long-context tasks such as book and audio summarization. 
\section{Conclusion}
\label{conclusion}
We introduced a new attention metric Fastmax, and showed that it has a wall-clock time and memory scaling that is linear in the number of tokens $N$, while having the same expressivity as vanilla attention (Softmax). Our approach stands out from previous methods attempting to create a subquadratic attention as a mechanism that does not compromise on sparsifying the attention matrix, localizing attention, or separating the local and far attention calculation processes under the assumption that far attentions have negligible values. Furthermore, Fastmax can be seamlessly integrated into any Transformer architecture by swapping out Softmax with Fastmax.
\par 
There are obvious next steps to try. The use of data-structures, as in \cite{Yang2003-mg}, might allow us to increase the order $p$, while dropping negligible terms. The use of custom gradients might allow a reduction in the complexity by a factor of $D$, as discussed in \S \ref{custom_gradients}. The algorithm may be suitable to try on CPU machines with large RAM to get past the lower RAM on fast GPUs (whose brute force computing power is needed to tackle the $O(N^2)$ complexity.

Having a linearly scaling attention metric enables Transformers to find new applications in domains such as audio, image, and video processing that Transformers were not usable before. A Transformer with quadratic time and memory complexity will always require compromises or expensive computing to process medium/long duration audio signals, high-resolution images/video frames, and larger language corpora. As a result, academic labs are reduced to of fine-tuning of pre-trained models, rather than training Transformers from scratch.

\subsection{Broader Impact}
\par Our work should have multiple positive impacts: including reducing energy consumption through making Transformers more efficient, making AI implementable on edge devices through reducing the computational complexity, and making large-scale AI much more accessible through the reduction of dependence on high-end GPUs. However, the increased accessibility of large-scale AI also raises concerns about potential negative societal impacts, such as the proliferation of deepfakes, which could be exploited for malicious purposes.

\section*{Acknowledgements}
\label{acknowledgements}
 Discussions over the last few months with several colleagues at UMD on the basics of transformers were useful. Ideas on high dimensional polynomials were developed during  previous discussions with (late) Dr. Nail Gumerov and Dr. Changjiang Yang. Partial support of ONR Award N00014-23-1-2086 is gratefully acknowledged.


\end{document}